# A Case Study on the Classification of Lost Circulation Events During Drilling using Machine Learning Techniques on an Imbalanced Large Dataset

Toluwalase A. Olukoga[1], Yin Feng[2]

University of Louisiana at Lafayette

**Abstract**

In this study, we present machine learning classification models that forecast and categorize lost circulation severity preemptively using a large class imbalanced drilling dataset with good prediction accuracy. We demonstrate reproducible core techniques involved in tackling a large drilling engineering challenge utilizing easily interpretable machine learning approaches.

We utilized a dataset of over 65,000 records with class imbalance problem from Azadegan oilfield formations in Iran. Eleven of the dataset's seventeen parameters are chosen to be used in the classification of five lost circulation events. To generate classification models, we used six basic machine learning algorithms and four ensemble learning methods. Linear Discriminant Analysis (LDA), Logistic Regression (LR), Support Vector Machines (SVM), Classification and Regression Trees (CART), k-Nearest Neighbors (KNN), and Gaussian Naive Bayes (GNB) are the six fundamental techniques. We also used bagging and boosting ensemble learning techniques in the investigation of solutions for improved predicting performance. The performance of these algorithms is measured using four metrics: accuracy, precision, recall, and F1-score. The F1-score weighted to represent the imbalance in the number of instances in the categories of drilling fluid lost circulation events in the dataset is chosen as the preferred evaluation criterion.

When the prediction performance of the six basic models was compared, the CART model was found to be the best in class for identifying drilling fluid circulation loss events with an average weighted F1-score of 0.9904 and standard deviation of 0.0015. Upon application of ensemble learning techniques, a Random Forest ensemble of decision trees showed the best predictive performance. It identified and classified lost circulation events with a perfect weighted F1-score of 1.0. Using Permutation Feature Importance (PFI), the measured depth was found to be the most influential factor in accurately recognizing lost circulation events while drilling.

The novelty of this work is the reproducibility of the workflow for implementing easily interpretable machine learning classification techniques for predicting lost circulation events. By following the workflow in this study, drilling teams will be able to better plan and conduct corrective actions before drilling fluid losses occur. This capability decreases drilling costs and accelerates resource recovery.

## 1 Introduction

When drilling in most formations, drilling fluid circulation loss is a significant issue. It can result in expensive and time-consuming rig downtime due to stuck pipes and well control events. As a result, being able to forecast this occurrence to anticipate and prevent it is critical for improving drilling success. Historically, empirical (Lavrov and Tronvoll, 2004; Majidi et al., 2010; Shahri et al., 2012) and machine learning techniques (Hosseini, 2017; Abbas et al., 2019; Agin et al., 2020) have been utilized to address this difficulty. Empirical methods, on the other hand, fall short of capturing the complexities of the relationship between independent factors and circulation loss events. While many machine learning applications to lost circulation severity categorization in the literature have limited predictive accuracy when applied to big datasets (Mardanirad et al., 2021). A previous work by Mardanirad et al. (2021) using a large drilling dataset to classify lost circulation events made use of deep neural networks that are not easily interpretable.

---

[1] tolu.olukoga@gmail.com    [2] yin.feng@louisiana.edu

The goal of this work is to forecast and categorize lost circulation severity using a large drilling dataset with good prediction accuracy. Clearly interpretable machine learning algorithms are applied in a reproducible manner to a drilling engineering case study. Most suitable machine learning classification techniques for this dataset is identified. As a result of this study's findings, drilling teams can better plan and execute remedial procedures before drilling fluid losses occur. The ability to do so reduces drilling costs and speeds up resource recovery.

We broadly classified the workflow followed in this study into four steps as shown in Figure 1.

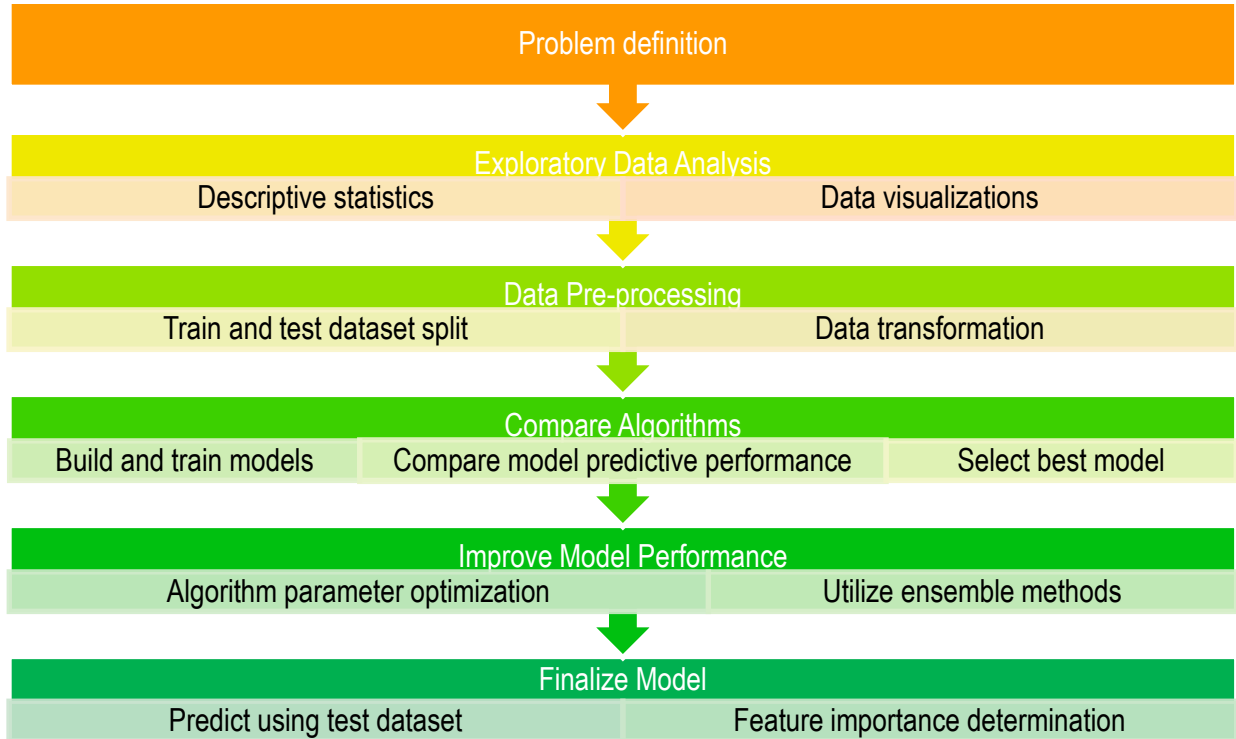

Figure 1. Workflow for comparing data-driven machine learning models

First, we define the problem we seek to solve in this study. Then we gather and explore the data through analysis and wrangling. Next, we compare a battery of basic machine learning algorithm to identify one that best models the data. Upon identification of a best-in-class algorithm, we improve the predictive performance of the algorithm for this dataset by tuning its parameters and employing ensemble learning techniques. Finally, the best model predicts the test dataset and the importance of the input features to the final model is determined.

## 2 Exploratory Data Analysis

The data set was evaluated to determine its structure, datatype, scale, feature distributions, and any correlations or outliers.

**2.1 Descriptive statistics.** This study makes use of a dataset provided by Mardanirad et al. (2021) generated from twenty wells drilled into Iran's Azadegan oilfield. The authors had removed all missing, duplicate, or inconsistent data from the dataset. The dataset comprises of 65,376 data records, each of which contains values for seventeen numerical drilling fluid and mechanical variables. Table 1 provides a statistical summary of the variables in the dataset in a descriptive manner. The variables served as input features in this analysis to estimate the severity of drilling fluid circulation loss incidents.

Table 1. The descriptive statistical breakdown of the dataset.

| Feature | Average | Std. Dev. | Minimum | 25% Percentile | 50% Percentile | 75% Percentile | Maximum |
|---|---|---|---|---|---|---|---|
| Hole section, inch | 11.61 | 4.51 | 4.125 | 8.5 | 8.5 | 17.5 | 26 |
| Measured depth, m | 2046 | 1048 | 14 | 1172 | 2098 | 2905 | 4285 |
| Rate of Penetration, m/hr | 11 | 14 | 0 | 4 | 6 | 12 | 616 |
| Weight on bit, klbs | 16 | 11 | 0 | 8 | 13 | 21 | 79 |
| Rotation, RPM | 161 | 49 | 0 | 130 | 155 | 186 | 457 |
| Torque, klbs-ft | 3 | 2 | 0 | 2 | 3 | 4 | 12 |
| Standpipe Pressure, psi | 2047 | 594 | 9 | 1661 | 2137 | 2542 | 3915 |
| Flow in, gpm | 593 | 222 | 0 | 478 | 593 | 782 | 1934 |
| Flow out, % | 60 | 16 | 0 | 50 | 58 | 72 | 100 |
| Pump stroke, spm | 112 | 37 | 6 | 95 | 115 | 141 | 242 |
| Mud weight (pcf) | 79 | 11 | 66 | 75 | 77 | 80 | 127 |
| Funnel viscosity (seq/qt) | 43 | 6 | 29 | 39 | 43 | 45 | 76 |
| Plastic viscosity (cp) | 14 | 7 | 3 | 10 | 13 | 16 | 48 |
| Yield point (lbs/100 ft$^2$) | 16 | 5 | 2 | 13 | 15 | 19 | 47 |
| Gel strength 10 s (lbs/100 ft$^2$) | 4 | 2 | 1 | 3 | 4 | 5 | 26 |
| Gel strength 10 min (lbs/100 ft$^2$) | 6 | 3 | 1 | 4 | 5 | 7 | 27 |
| Solid, % | 14 | 6 | 4 | 11 | 13 | 16 | 65 |

It was discovered that all the features in the dataset were numerical, with only the target feature (Loss Severity) being a categorical feature. The data in Table 1 is also observed to have different scales. Measured depth and standpipe pressure, for example, have values in the thousands, whereas torque, plastic viscosity, and gel strength have unit values. Most machine learning algorithms require similar-scale input features. This is done so that features with values in the thousands are not given more weight than those with values in units simply because of the magnitude of their values. To address this, transformations will be used to convert the dataset's values to the same scale. Section 4 contains more information about the transformation used in this study.

In the dataset, five distinct categories of drilling fluid circulation loss events associated with these variables were identified based on severity. These loss events are Complete Loss, Severe Loss, Partial Loss, Seepage Loss, and No Loss. Figure 2 depicts the proportion of data records in each category of circulation loss occurrence. It is discovered that the categories are imbalanced, with 76 percent of the data classified as No-Loss. The categories of total loss and severe loss account for less than 1% of the dataset.

To address the extreme degree of imbalance in the data set, special data or algorithmic solutions may be required. Data preprocessing techniques may require the use of more appropriate sampling approaches that over- or under-sample the dataset's classes (Chawla, 2009; Ganganwar, 2012; Ramyachitra and Manikandan, 2014). Cost-sensitive machine learning methods such as k-nearest neighbors (kNN), decision trees, neural networks, and modified support vector machines (SVMs) have also been suggested for dealing with imbalanced datasets with/without the data being specially sampled (Ganganwar, 2012).

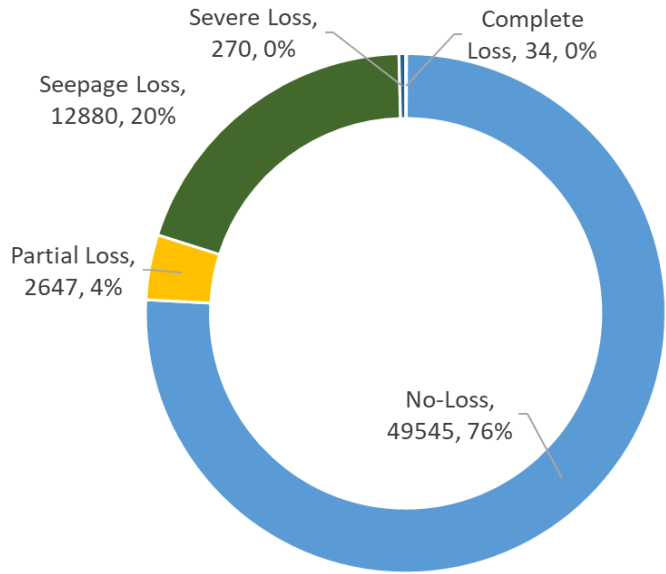

Figure 2. A breakdown of data records in each category of circulation loss occurrence

**2.2 Data visualization.** The data is then visualized using histograms (see Figure 3) to better understand the distribution of the data. Many of the features have skewed Gaussian/normal distributions. The underlying assumption for most machine learning algorithms is that input datasets have a normal distribution. As a result, transformations to correct the skewness and make the distribution of all features more normal may be required to prepare this dataset for modeling with these algorithms. Section 4 discusses how skewness is addressed in this study.

Figure 4 depicts a typical box and whiskers plot to highlight the details of such a plot while Figure 5 shows the box and whisker plots for visualizing the dispersion of values in the dataset in this study. The features are observed to have a varying spread around their mean, with the means of all the features not lining up together. This implies that standardization may be necessary prior to modeling. Outliers can also be found in all features except the Hole Section and Measured depth.

Finally, the Pearson's correlations between the features are calculated and visualized using a multimodal heatmap (see Figure 6). The heatmap is a cross plot of all features on the x-axis against all features on the y-axis. Larger numbers closer to green indicate higher correlation, whereas dark red colors show small values indicating lower correlation between corresponding features. The diagonal values represent each feature that is correlated with itself; they are all 1 and green. High correlation is observed between several features, including the 0.95 green value seen at the intersection of mud weight and solids, as well as the 0.92 value colored green at the intersection of Gel Strength 10 sec and Gel Strength 10 minutes.

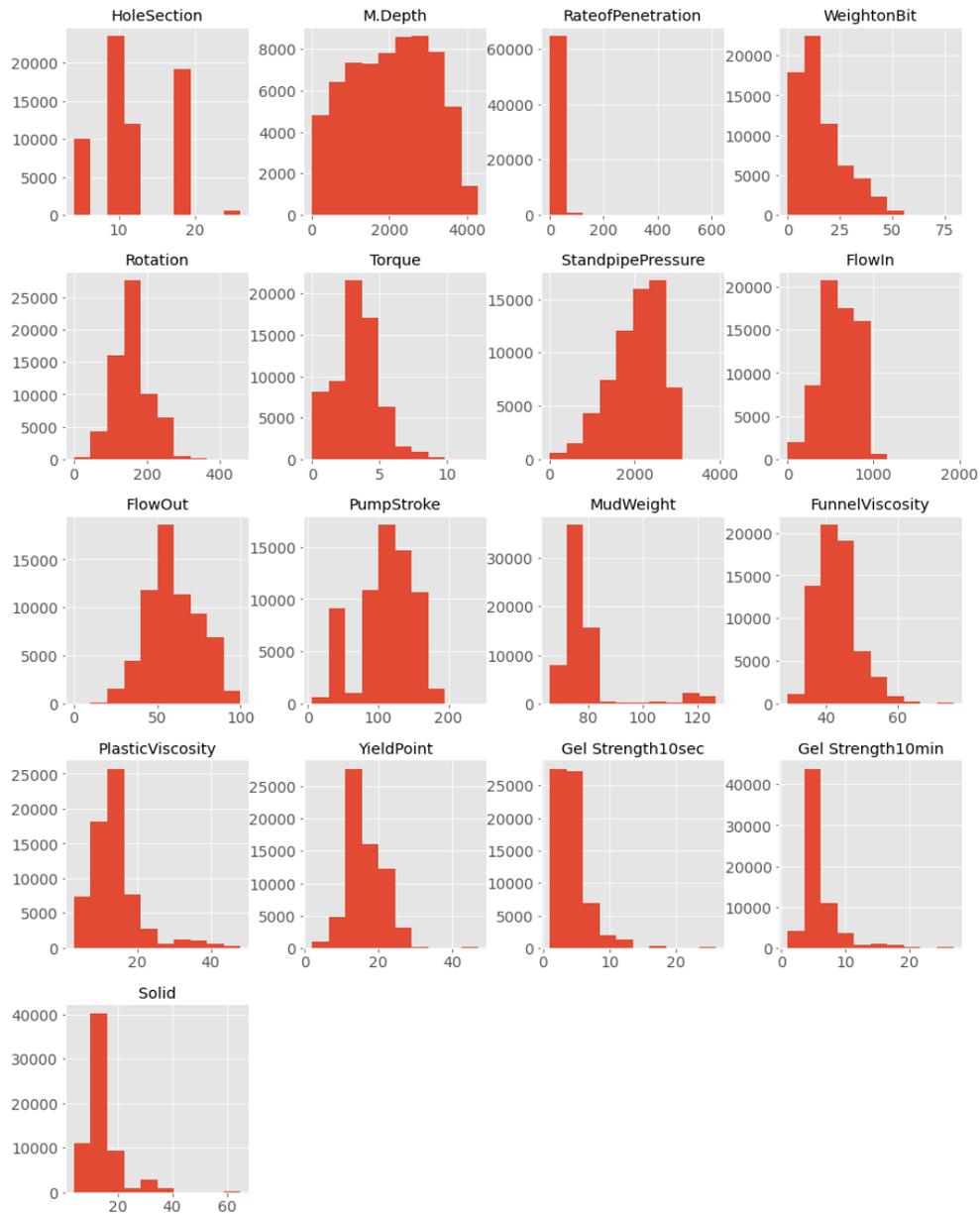

Figure 3. Representation of data set features as histogram plots

It is thought that eliminating features with a high correlation is important for model interpretability (Hall, 1999). If the primary goal of a study is prediction accuracy rather than model interpretability, removing highly correlated features is not considered important (Lieberman and Morris, 2014). Since both model prediction accuracy and interpretability are being investigated in this study, one of the pairs of highly correlated features is removed.

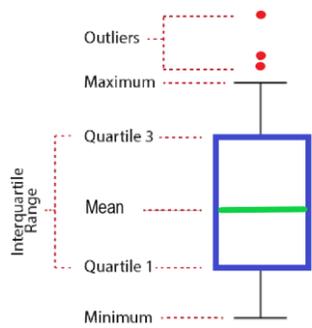

Figure 4. A typical box and whisker plot (Kreiley, 2020)

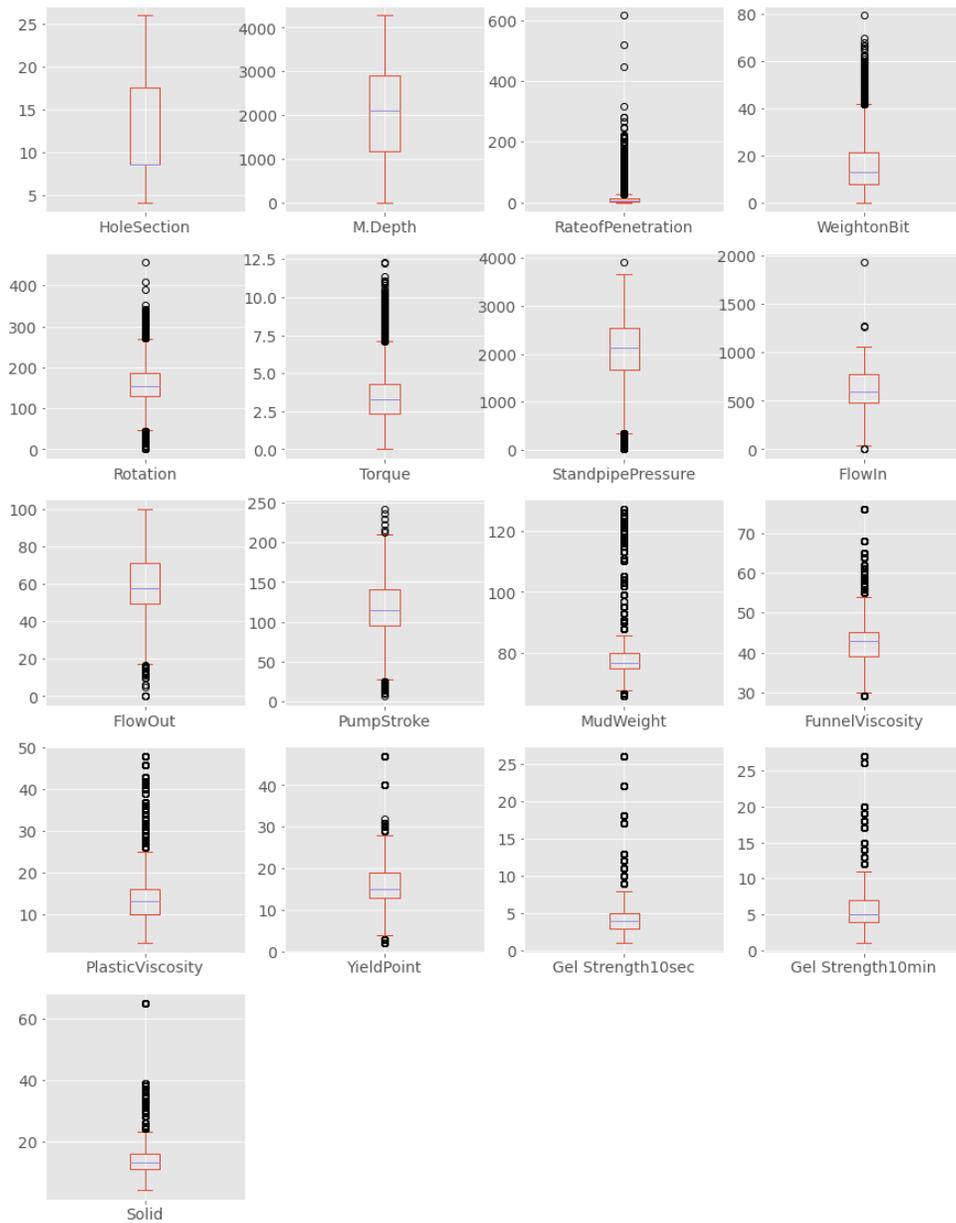

Figure 5. Representation of data set features as box and whisker plots

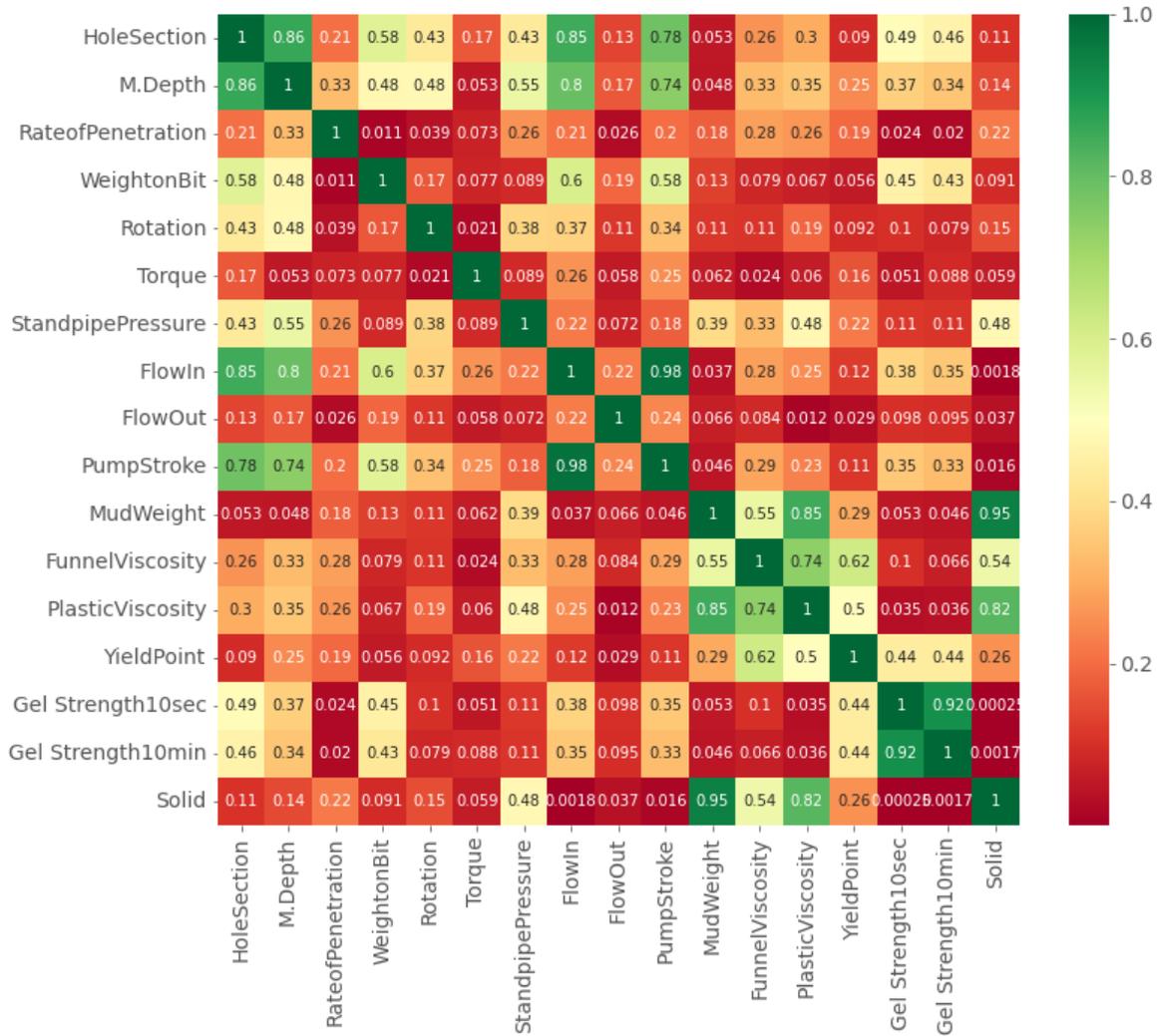

Figure 6. Visualization of Pearson's correlations between the dataset features using a multimodal heatmap

## 3 Data Pre-Processing

**3.1 Handle outliers**. To deal with outliers discovered during the data analysis stage, the first and third quartiles, as well as the interquartile range (IQR), for the variables in the dataset were calculated. By adding the third quartile values to 150 percent of the IQR, new maximum values were generated. In addition, new minimum values were calculated by removing 150 percent of the IQR from the first quantile values. Outlier values that exceeded the new maximum values were replaced with the new maximum values. Outlier values lower than the new minimum values, on the other hand, were substituted with the new minimum values.

**3.2 Feature selection.** In this study, high correlation was classified as features having Pearson's correlation values greater than 0.7. When two features are discovered to be highly correlated, one is chosen at random while the other is removed. Six features in this category were removed from the dataset. The eleven features chosen for the next steps were as follows:

- Measured depth, m
- Rate of Penetration, m/hr.
- Weight on bit, klbs
- Rotation, RPM
- Torque, klbs-ft
- Standpipe Pressure, psi
- Flow in, gpm
- Flow out, %
- Plastic viscosity (cp)
- Yield point (lbs./100 ft$^2$)
- Gel strength 10 min (lbs./100 ft$^2$)

Table 2 provides a descriptive statistical summary of the structure of the dataset utilized for modeling.

**3.3 Training and Test Dataset Split.** A validation hold-out (test) approach is used to validate created models and ensure that their performance is not confirmed using the same dataset that was used to train them. Using this method, the dataset was randomly mixed and separated into two representative subsets. Eighty percent of the data was used for modeling and analysis. The remaining 20% was kept aside as an unseen testing subset that was solely utilized to confirm the final model's classification accuracy.

Table 2. The descriptive statistical breakdown of the processed dataset.

| Feature | Average | Std. Dev. | Minimum | 25% Percentile | 50% Percentile | 75% Percentile | Maximum |
|---|---|---|---|---|---|---|---|
| Measured depth, m | 2046 | 1048 | 14 | 1172 | 2098 | 2905 | 4285 |
| Rate of Penetration, m/hr | 9.07 | 7.33 | 0.06 | 3.55 | 6.43 | 12.32 | 25.48 |
| Weight on bit, klbs | 15.59 | 10.7 | 0 | 7.5 | 12.7 | 21.2 | 41.75 |
| Rotation, RPM | 161 | 47 | 47 | 130 | 156 | 187 | 270 |
| Torque, klbs-ft | 3.26 | 1.66 | 0 | 2.37 | 3.32 | 4.28 | 7.14 |
| Standpipe Pressure, psi | 2070 | 593 | 340 | 1674 | 2147 | 2552 | 3864 |
| Flow in, gpm | 593 | 222 | 37 | 478 | 593 | 782 | 1238 |
| Flow out, % | 60 | 16 | 17 | 50 | 58 | 72 | 105 |
| Plastic viscosity (cp) | 13 | 5 | 3 | 10 | 13 | 16 | 25 |
| Yield point (lbs/100 ft2) | 16 | 5 | 5 | 13 | 15 | 20 | 28 |
| Gel strength 10 min (lbs/100 ft2) | 5.7 | 2.06 | 1 | 4 | 5 | 7 | 11.5 |

**3.4 Data Transformation.** The varying scale of the dataset, as discovered during the exploratory data analysis step, necessitates some type of data transformation to ensure optimal performance of specific algorithms. To avoid data leakage, machine learning pipelines were used to transform and build the models in the same processing step.

The values of all features were processed using the normalization approach described in Equation (1) to ensure all the features were scaled. The method involves re-scaling input variables to be between a range of 0 and 1. The minimum and maximum values apply to each variable to which the *x* value that is being normalized belongs.

$$y = \frac{x - x_{min}}{x_{max} - x_{min}} \quad \ldots\ldots\ldots\ldots(1)$$

Next, the skewness identified when the data was explored was addressed by using box-cox power transform. This transform stabilizes the variance of the distribution in the dataset thereby removing skewness. The transform makes the probability distributions of the features more normal/Gaussian. Figure 7 shows the histogram plots of the data after these transformation steps.

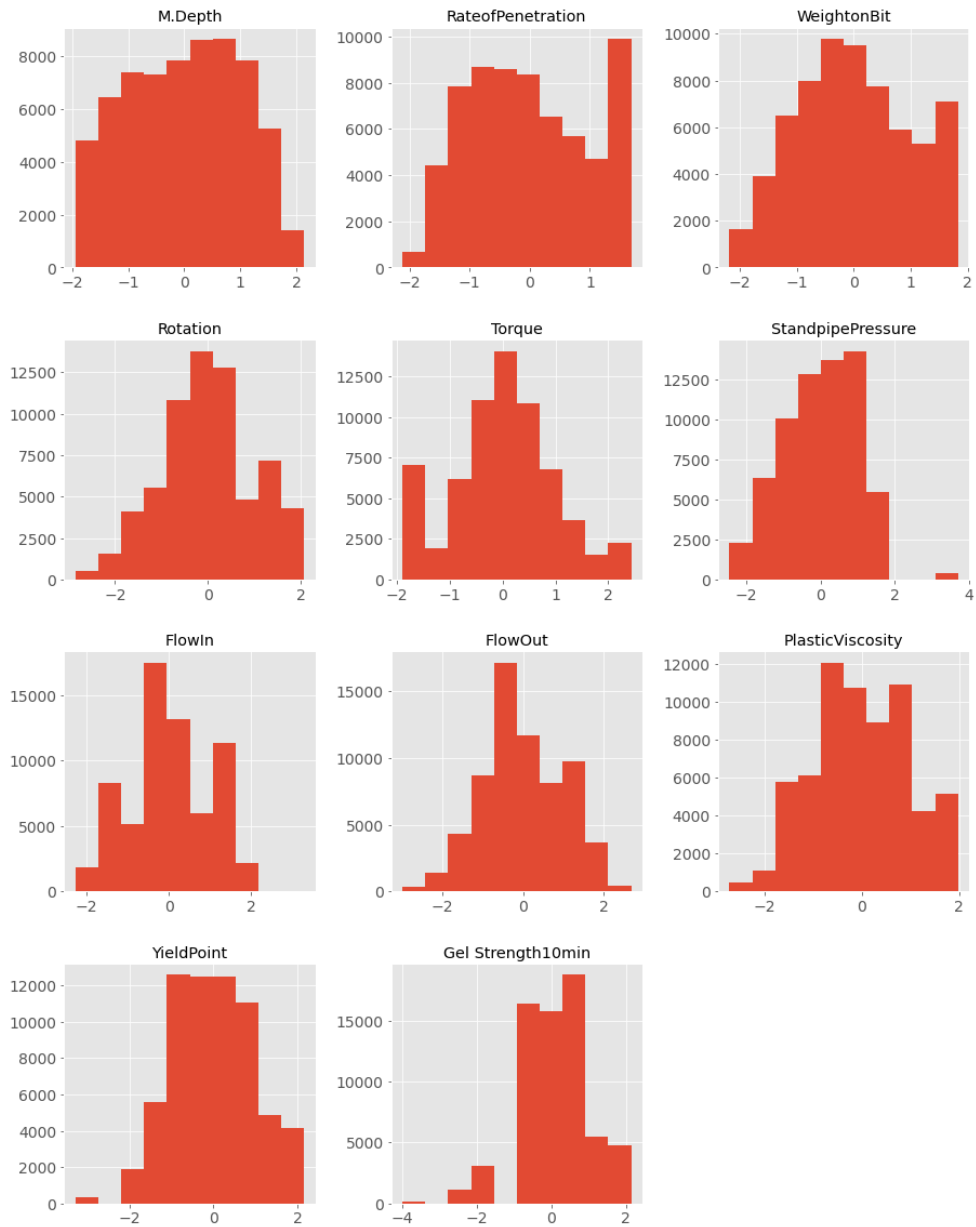

Figure 7. Representation of data set features as histogram plots after data preprocessing steps

## 4 Compare Algorithms

In this section we compare the classification performance of various machine learning algorithms on the lost circulation dataset.

**4.1 Build and train models.** In this study, we elect to handle the imbalance in the dataset by choosing an appropriate algorithm rather than through special data sampling techniques. From the exploratory data analysis step, cost-sensitive and distance-based algorithms such as k-nearest neighbors (kNN), decision trees, neural networks, and modified support vector machines (SVMs) are suggested as likely to have better performance. We therefore investigate some of these and other baseline algorithms to evaluate and identify the one that will be best for this data set. We evaluate linear and non-linear algorithms. The linear algorithms are Linear Discriminant Analysis (LDA) and Logistic Regression (LR). Support Vector Machines (SVM), Classification and Regression Trees (CART), k-Nearest Neighbors (KNN) and Gaussian Naive Bayes (NB) are the non-linear algorithms considered. These algorithms are built with the scikit-learn toolkit, which is basically a library for implementing, tweaking parameters, and evaluating machine learning algorithms, among other things (Pedregosa et al., 2011).

**4.2 Model evaluation criteria.** In this study, the predictive performance of the models is assessed using four criteria: accuracy, weighted-average precision, weighted-average recall, and the weighted-average F1-score. Equations 1- 4 show the equations for these metrics (Behera et. al, 2019). Confusion matrices that show the performance of models for each class of the target variable is also used.

$$Accuracy = \frac{\sum_{i=1}^{n}\frac{TP+TN}{TP+FP+TN+FN}}{n} \quad \dots\dots\dots\dots\dots(1)$$

$$Weighted\ Averaged\ Precision = \frac{\sum_{i=1}^{n}|s_i|\frac{TP_i}{TP_i+FP_i}}{\sum_{i=1}^{n}|s_i|} \quad \dots\dots\dots\dots(2)$$

$$Weighted\ Averaged\ Recall = \frac{\sum_{i=1}^{n}|s_i|\frac{TP_i}{TP_i+FN_i}}{\sum_{i=1}^{n}|s_i|} \quad \dots\dots\dots\dots(3)$$

$$Weighted\ Averaged\ F1-Score = \frac{\sum_{i=1}^{n}|s_i|2\frac{2TP_i}{TP_i+FP_i+FN_i}}{\sum_{i=1}^{n}|s_i|} \quad \dots\dots\dots\dots(4)$$

In the equations, the terms *TP*, *TF*, *FP*, and *FN* refer to the number of predicted value outcomes versus actual values. True positive (*TP*) is the number of actual positive values predicted to be positive, whereas true negative (*TN*) denotes the number of actual negative values predicted to be negative. The number of actual positive and negative values incorrectly classified by the model is represented by *FP* and *FN*, which stand for false positive and false negative.

The weighted average of the metrics is a more equitable metric for dealing with class imbalance in the dataset. The weighted average of all metrics is generated by averaging all per-class metric and factoring in the support for each class. *n* in the equations is the number of classes while *s* is the number of instances of a class in the dataset referred to as support.

While accuracy can be thought of intuitively as the ratio of all accurately predicted values to total values. Precision and recall can be better described by utilizing the example case of five known instances of lost circulation events across the period shown in Figure 8a. A classifier with perfect precision ($\frac{5}{5+0} = 1$) and perfect recall ($\frac{5}{5+0} = 1$) detects all these events at their known time of occurrence (Figure 8b). Figure 8c depicts a possible scenario with a classifier that detects exactly two of these events. Since all recognized events are true lost circulation events (True positives), this classifier has perfect precision ($\frac{2}{2+0} = 1$). However, because the classifier could not discover all actual occurrences, it has three false negatives and thus a low recall ($\frac{2}{2+3} = 0.4$). Figure 8d depicts another example in which the classifier detects all 5 true positive events but also classifies three other events as lost circulation events. In this case, the classifier has perfect recall ($\frac{5}{5+0} = 1$) but low precision ($\frac{5}{5+3} = 0.625$).

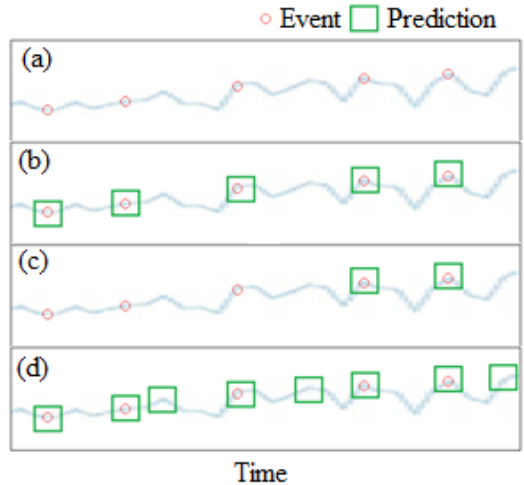

Figure 8. Depiction of Precision and Recall

Some scenarios require high precision while others require high recall. In situations when a balance of both is required the F1-score is preferred. The F1-score or F-measure represents the harmonic average of precision and recall. When precision and recall are equally important as is the case in this study, the F1-score is the preferred metric.

Also, we rely more on the weighted F1-score metric due to the general imbalance in the dataset. All the metrics can have values ranging between 0 and 1. When comparing results of models, values closer to 1 suggest a strong classifier.

**4.3 Compare model predictive performances.** With the baseline algorithms set to their default parameter settings, models of the dataset were built. 10-fold cross-validation was performed on the training dataset as a validation tool for all models. In 10-fold cross-validation, the training data set is divided into ten groups of about equivalent size, with one of the groups repeatedly put aside to test the model after it has been trained on the other nine groups (Sugiyama, 2016).

The performances of the models resulting from the six algorithms were compared using their average f1-scores. Figure 9 depicts, using box and whisker plots, the distribution of f1-score values calculated throughout cross validation folds for the six assessed models. A tight distribution is observed across all models and most especially with the CART and KNN suggesting low variance. The KNN and CART models have the highest F1-scores.

The CART model is observed to be the best performing model when measured by average weighted f1-score of 0.9904 and standard deviation of 0.0015. The KNN model is a strong contender with average f1-score of 0.9832 and standard deviation of 0.0016. The NB model had the worst performance with an average f1-score of 0.6921and standard deviation of 0.0052.

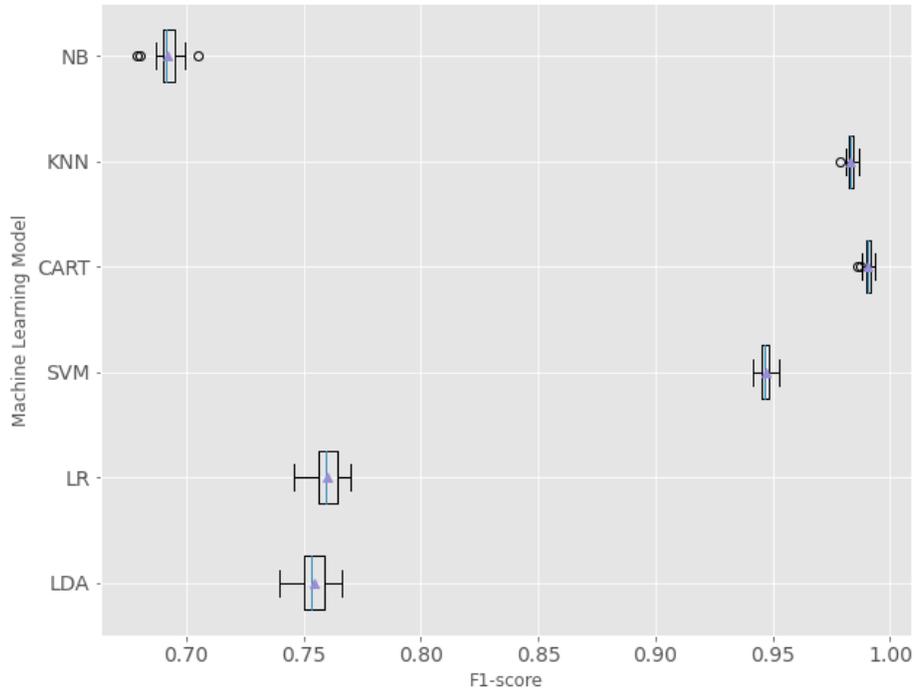

Figure 9. Box and whisker plots depicting the distribution of f1-score values generated over cross validation folds for the six evaluated models.

**4.4 Results of best performing model.** Table 3 displays the results of the four-assessment metrics (precision, recall, f1-score, and accuracy) for all target variable classes for the CART model when predicting test dataset. In Table 3, support is defined as the number of cases that supported the prediction of that class. The weighted average f1-score is determined to be 1.0, indicating that the model performs well in categorizing previously unseen data. When it comes to classifying Complete Loss, the model has the worst predictive performance. The model only has 8 examples for its predictions of this class. The aggregate weighted f1-score for all target variable classes was 0.99. On the classification of the Complete Loss events, the model obtained the lowest score across all measures.

Figure 10 shows a more detailed representation of the CART model's classification for each target variable class (No Loss, Seepage Loss, Partial Loss, Severe Loss, and Complete Loss). The figure depicts a confusion matrix of predicted and actual values for each class in the test dataset. Along the matrix's main diagonal, the number of true positives for each class is displayed, while the numbers on each row represent the number of occurrences misclassified as that class.

The CART model accurately identified 99.48 percent of the 9,860 true cases of the No Loss class, with misclassified instances across all classes. The Seepage Loss class is misclassified by the model 46 times out of 2,566 times. The model accurately recognized 559 and 68 of the 572 and 70 cases in the Partial Loss and Severe Loss classes, respectively. Finally, the CART model properly identifies 7 of the 8 Complete Loss events, with one instance mislabeled as a Partial Loss event. Overall, the model did a good job of correctly identifying the lost circulation events in the dataset.

Table 3. Classification report for the CART model

|  | Precision | Recall | F1-score | Support |
|---|---|---|---|---|
| **No Loss** | 1 | 0.99 | 1 | 9860 |
| **Seepage Loss** | 0.98 | 0.98 | 0.98 | 2566 |
| **Partial Loss** | 0.96 | 0.98 | 0.97 | 572 |
| **Severe Loss** | 0.96 | 0.97 | 0.96 | 70 |
| **Complete Loss** | 0.70 | 0.88 | 0.78 | 8 |
| **Macro average** | 0.92 | 0.96 | 0.94 | 13076 |
| **Weighted average** | 0.99 | 0.99 | 0.99 | 13076 |
| **Accuracy** |  | 0.99 |  | 13076 |

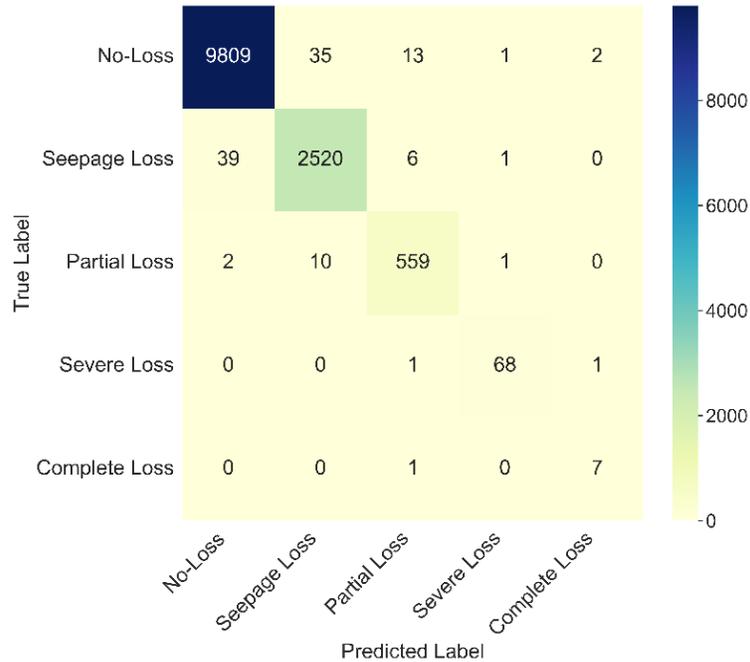

Figure 10. A confusion matrix for CART model on test dataset

**5 Improve Model Classification Performance**

More research was conducted to improve the CART model's performance in classifying the severity of lost circulation events. First, the effect of tuning the model's parameters is examined and then application of ensemble learning techniques are investigated.

**5.1 Algorithm parameter optimization.** Since the CART model was built using default parameters (no specified maximum depth and minimum sample leaf of one among other setting), the first step in improving model performance is to tune the parameters of the model. GridSearchCV, a scikit-learn library method for identifying ideal values for specified parameters that minimize any preferred model metric, was utilized (Feurer and Hutter, 2019). As input, the method generates and evaluates one model for each parameter combination. Each model was evaluated using a 10-fold cross-validation. When several parameters such as maximum depth, maximum features, minimum samples split, and minimum sample leaf were all tuned, the predictive performance of the CART model was observed to deteriorate drastically. Therefore, only the maximum depth allowed for the decision tree was tuned. A model with maximum depth of 53 performed the best.

The optimized model was then utilized to classify the test dataset's rows/instances. Table 4 shows the classification report for the model. The confusion matrices obtained when the CART model was evaluated with unseen test dataset is shown in Figure 11. It was observed that parameter tuning had no effect on the performance of the CART model. The classification report and confusion matrix show the exact values obtained for the untuned model on all evaluation metrics.

Table 4. Classification report for the tuned CART model

|  | Precision | Recall | F1-score | Support |
|---|---|---|---|---|
| **No Loss** | 1 | 0.99 | 1 | 9860 |
| **Seepage Loss** | 0.98 | 0.98 | 0.98 | 2566 |
| **Partial Loss** | 0.96 | 0.98 | 0.97 | 572 |
| **Severe Loss** | 0.96 | 0.97 | 0.96 | 70 |
| **Complete Loss** | 0.70 | 0.88 | 0.78 | 8 |
| Macro average | 0.92 | 0.96 | 0.94 | 13076 |
| Weighted average | 0.99 | 0.99 | 0.99 | 13076 |
| Accuracy |  | 0.99 |  | 13076 |

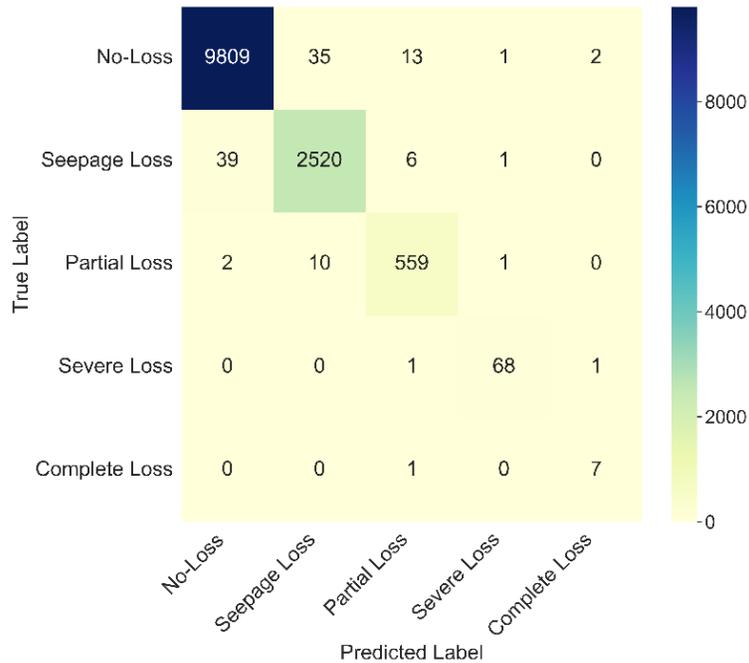

Figure 11. A confusion matrix for the tuned CART model on test dataset

**5.2 Ensemble methods.** Another investigation is carried out to potentially obtain a model with better performance by employing two of the most common ensembles learning approaches – Bagging and Boosting.

Using the bagging ensemble learning technique, numerous instances of models generated using an algorithm are independently fitted using subsamples (usually bootstrap sampled) of the dataset and utilized to make a prediction. The forecasts from each model instance are then combined to produce a final prediction for the Bagged meta-model.

The boosting strategy follows a similar procedure to the bagging method, except that the first model instance predicts using the entire dataset and the subsequent models attempt to rectify the errors of the previous model instances.

The optimized CART model from the preceding section is utilized as a base estimator for *Ada Boost Classifier* (ABC) and *Bagging Classifier (BC),* two Scikit-learn classifier classes that allow most algorithms to be used as their underlying estimator. Furthermore, the popular bagging and boosting ensemble algorithms, *Random Forest (RFC),* and *Gradient Boosting (GBC)*, are employed.

*5.2.1 Compare predictive performances of ensemble models.* The default parameter settings for these four ensemble algorithms are maintained as their performances in modeling the dataset on loss circulation are compared.

The performances of the models resulting from the four algorithms were compared using their weighted f1-scores. 10-fold cross validation was employed and the average of the weighted f1-scores was recorded for each model. The distribution of the average weighted f1-score values generated through cross validation folds for the evaluated ensemble models is depicted in Figure 12 using box and whisker plots.

While all ensemble models achieved high weighted f1-score, the Random Forest classifier model is shown to be the highest performing model, with an average weighted f1-score of 0.9965. With an average f1-score of 0.9956, the Ada Boosted classifier model comes in quite close in performance. The Gradient Boosting Classifier model performed the worst, with an average f1-score of 0.9344.

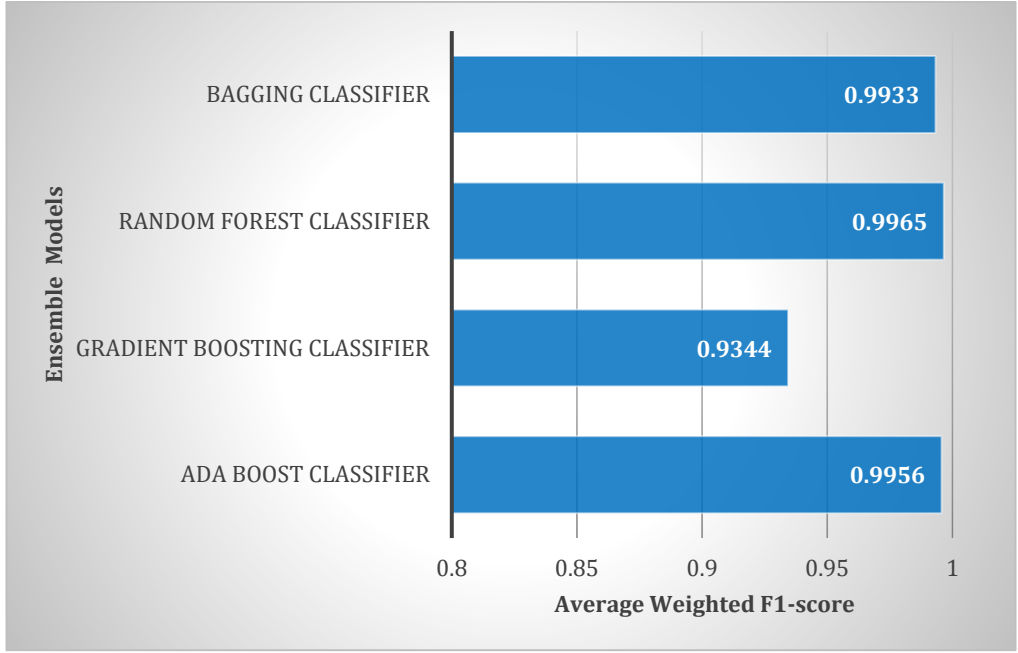

Figure 12. Average weighted F1-score values generated over cross validation folds for the four ensemble models.

*5.2.2 Ensemble algorithm parameter optimization.* A grid search of the optimal parameters for the Random Forest ensemble model showed that 156 base estimators, and maximum depth of seventeen with samples bootstrapped performed the best.

Table 5 shows the classification report with the four-evaluation metrics when the ensemble learning model was used to predict the test dataset. These results show that the optimized Random Forest ensemble model performed better than the optimized CART model in categorizing lost circulation events during drilling. The weighted F1-score was 1.00. which was an improvement on the 0.99 obtained with the tuned CART model (see Table 4). The Random Forest model got 0.82 F1-score on its highest misclassified class, Complete Loss. This contrasts with the 0.78 obtained by the CART model.

Figure 13 shows the confusion matrix of predicted and actual values for each class in the test dataset. Except for three instances of the most populous No Loss class, the Random Forest ensemble model properly classifies them all. The Seepage Loss class misclassifies 14 of the 2566 instances. Four and three of the 572 instances with the correct label of Partial Loss were misclassified as No Loss and Seepage Loss, respectively. The least populous categories, Severe Loss and Complete Loss, had 2 and 1 misclassified instances, respectively, in the test dataset. Overall, the Random Forest ensemble model works well in recognizing and classifying lost circulation events during drilling given the big dataset.

Table 5. Classification report for the Random Forest ensemble model

| | Precision | Recall | F1-score | Support |
|---|---|---|---|---|
| **No Loss** | 1 | 1 | 1 | 9860 |
| **Seepage Loss** | 1 | 0.99 | 1 | 2566 |
| **Partial Loss** | 0.99 | 0.99 | 0.99 | 572 |
| **Severe Loss** | 1 | 0.97 | 0.99 | 70 |
| **Complete Loss** | 0.78 | 0.88 | 0.82 | 8 |
| **Macro average** | 0.95 | 0.97 | 0.96 | 13076 |
| **Weighted average** | 1.00 | 1.00 | 1.00 | 13076 |
| **Accuracy** | | 1.00 | | 13076 |

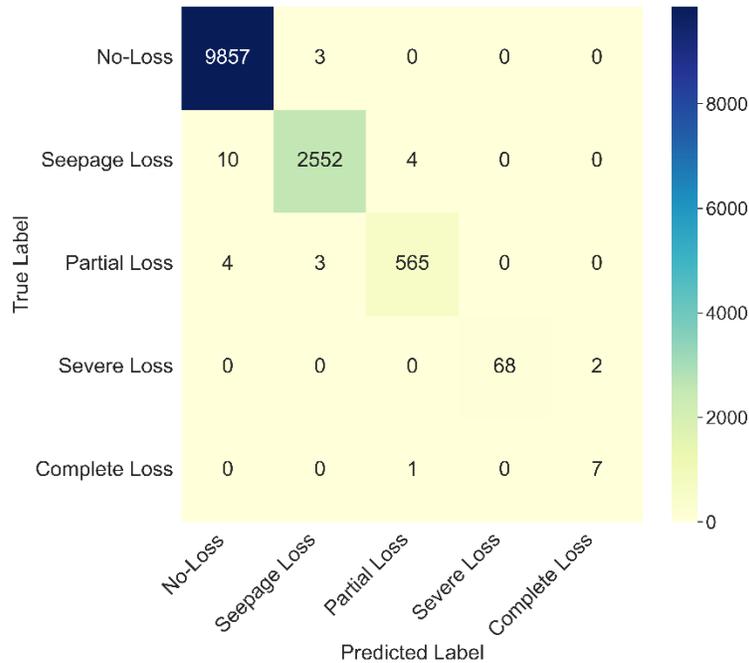

Figure 13. A Confusion Matrix for the Random Forest Ensemble Model

**5.3 Comparison to Previous work.** Mardanirad et al. (2021) employed a Convolutional Neural Network with 6 layers in their investigation with the same dataset. A 64-filter convolution layer, a max-pooling layer, one flat and one dropout layer, and two dense layers were among the layers. Table 6 and Figure 14 show the classification report and confusion matrix after they evaluated the classifier on 25% of the 65,376 data records.

Table 6. Classification report for Convolutional Neural Network model (after Mardanirad et al., 2021)

|  | **Precision** | **Recall** | **F1-score** | **Support** |
|---|---|---|---|---|
| **No Loss** | 0.99 | 0.98 | 0.99 | 12403 |
| **Seepage Loss** | 0.94 | 0.90 | 0.92 | 3191 |
| **Partial Loss** | 0.75 | 0.95 | 0.84 | 665 |
| **Severe Loss** | 0.94 | 0.90 | 0.92 | 76 |
| **Complete Loss** | 0.88 | 0.70 | 0.78 | 9 |
| **Accuracy** | | <1.00 | | 13076 |

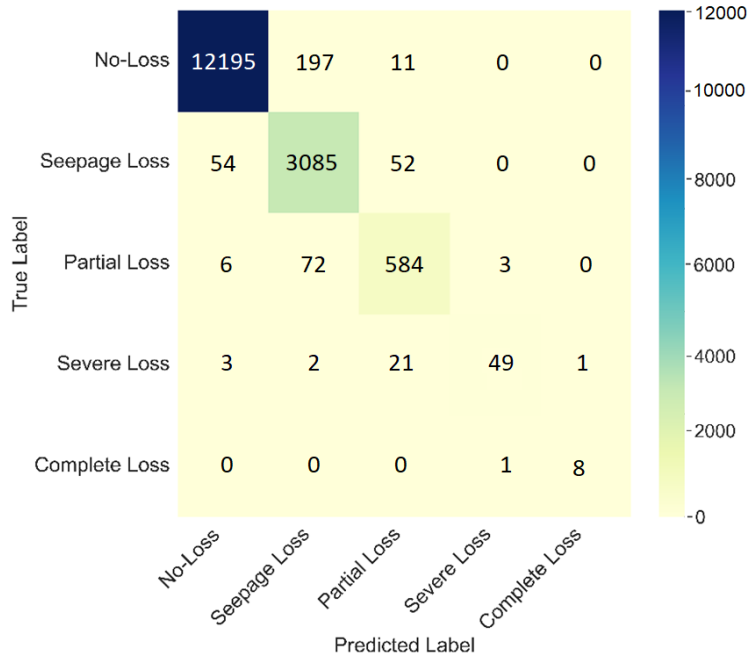

Figure 14. A Confusion Matrix for Convolutional Neural Network model (after Mardanirad et al., 2021)

While there are differences in preprocessing and data treatment between our study and the work of Mardanirad et al. (2021), the results are comparable. Furthermore, the Random Forest ensemble model, the best model in this study, offers the advantage of model interpretability without compromising model accuracy. This is covered in the following section.

## 6 Feature Importance

The features that were regarded to be significant in the prediction accuracy for the Random Forest ensemble model were explored. The permutation feature importance (PFI) metric, which evaluates the increase in model prediction error when feature values are randomly combined, is used (Altmann et al., 2010). Since features are assumed to have been chosen to ensure the minimal amount of error, the least amount of prediction error is obtained with all features; thus, any feature removal will always have a negative impact on model performance.

In this investigation, the highest possible weighted F1-score is obtained when the model is trained with all eleven input variables. As a result, the Random Forest ensemble model is developed and trained using the eleven variables, and the resultant weighted F1-score is used as the baseline F1-score. The model is then trained by eliminating one of the features iteratively, with the weighted F1-score on classification recorded. The increase in model error which is the difference between the baseline and weighted F1-scores when a feature is eliminated represents the permutation importance of that feature (Olukoga and Feng, 2021). Figure 15 depicts a permutation significance plot for the Random Forest ensemble model.

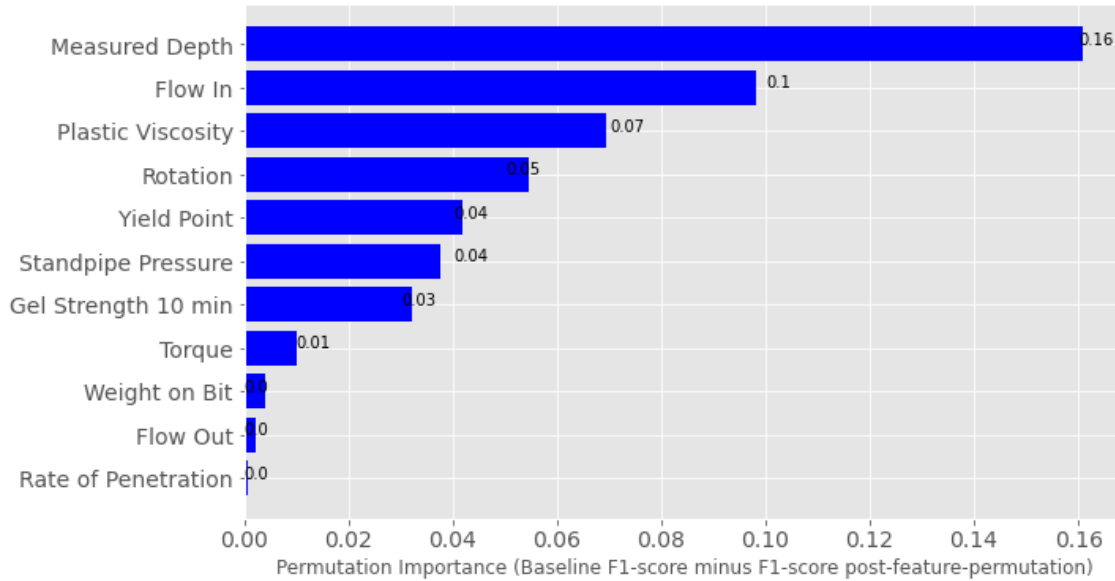

Figure 15. Permutation Significance Plot for the Random Forest ensemble model

The most significant factor in effectively categorizing lost circulation occurrences during drilling is discovered to be measured depth. Its removal from predictive evaluation reduces the weighted F1-score from 1.0 by 0.16 to 0.84. Volume of flow into the well and plastic viscosity were the second and third most important features, with 0.1 and 0.07 weighted F1-score decreases, respectively. Drill string rotation is found to be more critical than penetration rate.

The identification of the Random Forest ensemble model's most crucial features facilitates interpretation of its predictions. This is a significant benefit over black box approaches such as neural networks.

**7 Conclusion**

In this work, a workflow using readily interpretable machine learning approaches in a reproducible manner were presented for classifying the severity of circulation loss occurrences while drilling. A huge, preprocessed dataset of over 65,000 records from Azadegan oilfield formations in Iran is utilized. Eleven of the dataset's seventeen parameters are chosen to be used in the classification of five lost circulation events.

To generate classification models, six basic machine learning algorithms and four ensemble learning methods were used. Linear Discriminant Analysis (LDA), Logistic Regression (LR), Support Vector Machines (SVM), Classification and Regression Trees (CART), k-Nearest Neighbors (KNN), and Gaussian Naive Bayes (GNB) are the six fundamental techniques (NB). Bagging and boosting ensemble learning techniques were also used in the investigation of solutions for improved predicting performance.

The performance of these algorithms is assessed using four assessment metrics: Accuracy and weighted-average of Precision, Recall, and F1-score. The average f1-score weighted to represent the imbalance in the number of instances in the categories of drilling fluid circulation loss events in the dataset is chosen as the preferred evaluation criteria. The study draws the following conclusions:

- When the prediction performance of the six models was compared, the CART model was found to be the best in class for identifying drilling fluid circulation loss events with an average weighted f1-score of 0.9904 and standard deviation of 0.0015
- The CART model's performance did not improve because of parameter adjustments.
- The two models created by employing the CART model as base estimators for a bagging and boosting ensemble learning technique, a Random Forest bagging ensemble model, and a Gradient Boosting ensemble model, all achieved a high weighted f1-score.
- The Random Forest ensemble of decision trees showed the best predictive performance in finding and classifying lost circulation events, with a perfect weighted f1-score of 1.0.
- The influential parameters in accurately recognizing lost circulation events while drilling with the Random Forest ensemble model are ranked from greatest to least as follows: Measured depth - Flow into of well - Plastic viscosity – Rotation - Yield point - Standpipe Pressure - Gel strength 10 min – Torque - Weight on bit - Flow out of well - Rate of Penetration

By following the workflow in this study, drilling teams will be able to better plan and conduct corrective actions before drilling fluid losses occur. This capability decreases drilling costs and accelerates resource recovery.